\documentclass[a4paper, 10pt, conference]{ieeeconf}      
\usepackage{FG2017}
\usepackage{graphicx}
\usepackage{epsfig}
\usepackage{amsmath}
\usepackage{amsfonts}
\usepackage{emptypage}
\usepackage{epstopdf}
\usepackage{url}
\usepackage{caption}
\usepackage{subcaption}
\newcommand*{\affmark}[1][*]{\textsuperscript{#1}}

\FGfinalcopy 

\IEEEoverridecommandlockouts                              
\title{\LARGE \bf
Automatic Detection of ADHD and ASD from Expressive Behaviour in RGBD Data}

\author{\parbox{16cm}{\centering
    {\large Shashank Jaiswal\affmark[1] \hspace{0.5cm} Michel F. Valstar\affmark[1] \hspace{0.5cm} Alinda Gillott\affmark[2] \hspace{0.5cm} David Daley\affmark[3]}\\
    {\normalsize \affmark[1]School of Computer Science, The University of Nottingham}\\
		{\normalsize \affmark[2]Nottingham City Asperger Service \& ADHD Clinic}\\
		{\normalsize \affmark[3]Institute of Mental Health, The University of Nottingham}}
}

\begin{document}

\ifFGfinal
\thispagestyle{empty}
\pagestyle{empty}
\else
\author{Anonymous FG 2017 submission\\-- DO NOT DISTRIBUTE --\\}
\pagestyle{plain}
\fi
\maketitle

\begin{abstract}

Attention Deficit Hyperactivity Disorder (ADHD) and Autism Spectrum Disorder (ASD) are neurodevelopmental conditions which impact on a significant number of children and adults. Currently, the diagnosis of such disorders is done by experts who employ standard questionnaires and look for certain behavioural markers through manual observation.  Such methods for their diagnosis are not only subjective, difficult to repeat, and costly but also extremely time consuming. In this work, we present a novel methodology to aid diagnostic predictions about the presence/absence of ADHD and ASD by automatic visual analysis of a person’s behaviour. To do so, we conduct the questionnaires in a computer-mediated way while recording participants with modern RGBD (Colour+Depth) sensors. In contrast to previous automatic approaches which have focussed only detecting certain behavioural markers, our approach provides a fully automatic end-to-end system for directly predicting ADHD and ASD in adults. Using state of the art facial expression analysis based on Dynamic Deep Learning and 3D analysis of behaviour, we attain classification rates of 96\% for Controls vs Condition (ADHD/ASD) group and 94\% for Comorbid (ADHD+ASD) vs ASD only group. We show that our system is a potentially useful time saving contribution to the diagnostic field of ADHD and ASD.

\end{abstract}

\section{INTRODUCTION}

\noindent The last 5 years have seen a steady progress in automatic expressive behaviour analysis, with the detection and tracking of faces \cite{zhu2012}\cite{mathias2014}\cite{Xiong13-SDM}, recognition of facial muscle actions \cite{valstar2012fully}\cite{chu2013}\cite{ValstarEtAl2015_FER}, and accurate head pose estimation \cite{zhu2012}\cite{yan2016} all now possible under mild environmental constraints. This has renewed the interest of researchers to employ such behaviour analysis in the medical domain, targeting so-called \emph{behaviomedical} conditions that alter one's expressive behaviour \cite{Valstar2014_ABU}. In this paper we use state of the art facial expression analysis and RGBD head motion analysis to help in the diagnosis of Attention Deficit Hyperactivity Disorder (ADHD) and Autism Spectrum Disorder (ASD).

ADHD is a neurodevelopmental condition affecting a large number of people and it has been estimated that at least 2.5\% of the general adult population is affected by it \cite{simon2009}. ADHD is characterized by symptoms such as hyperactivity, impulsivity, inattention, etc. \cite{barkley1997,spencer2007}. It usually begins in early childhood and quite often the symptoms persists into adulthood \cite{weiss1985}. It is widely believed that both genetic \cite{stevenson2005} and environmental influences \cite{larsson2004} contribute to the underlying cause of this disorder. Presently, the diagnosis of ADHD is made following the criteria of the DSM-5 \cite{dsm}, which involve mechanisms to validate hyperactivity, attention deficit and impulsivity. The diagnosis is made by experts using a combination of developmental history, collateral information, psychometrics and behavioural observation and impairment. This is often difficult and time consuming.

\begin{figure}[t]
\begin{center}
  \includegraphics[width=0.9\linewidth]{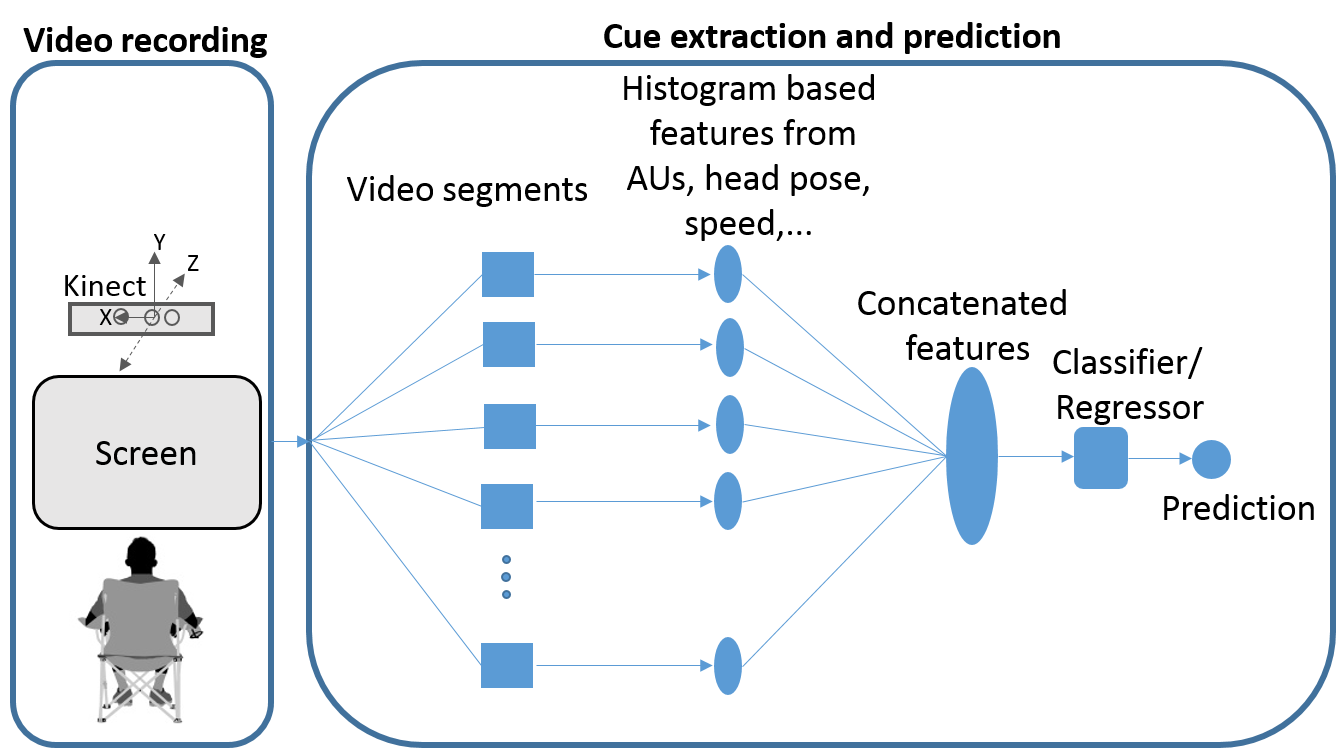}
\end{center}
  \caption{Overview of our system. A participant follows instructions on a screen while being recorded by a Kinect 2 camera. Deep Learning and RGB-D behaviour analysis of each video segment leads to successful ASD/ADHD classification.}
\label{graphical_abstract}
\end{figure}

ADHD is also known to show co-morbidity with ASD (Autism Spectrum Disorder). ASD is a neuro-developmental condition which is characterised by impairments in social interaction and communication and restricted, repetitive or stereotyped behaviours and interests. It has been found that a significant number of people with ASD also show symptoms of ADHD \cite{rao2013}. Treatment methods also vary for all 3 groups of people i.e. only ASD, only ADHD, and ADHD+ASD. Hence, accurate diagnosis can have important implications for treatment. However, currently the manual diagnosis for each of these disorders has to be done separately which requires more time.

Although there has been a lot of research in the area of ADHD and ASD and their diagnosis using brain scanners and manual observation of subjects for extended period of time by psychological experts, there has been relatively much less work in the direction of developing automated diagnostic tools for ADHD and ASD using easily available devices (e.g. video camera). The current methods of diagnosis are not only time consuming but they are also susceptible to human decision making bias. Development of machine learning methods which can be used as a  tool for decision making by human experts, could not only save time but will also help in bringing more objective, repeatable measures in the decision making process.

Currently available commercial systems (e.g. QbTest \cite{qb})  seek to automate the process of ADHD diagnosis uses only head motion of a person as a proxy for the activity of the subject. The other aspects of head actions including its pose are not taken into account directly. The head motion itself is captured using a normal 2D imaging camera which has limited ability to capture motion in 3D. Facial expression is another aspect which is completely ignored in  current ADHD assessment systems. Facial expressions and gestures can provide important cues about the psychological state of a person. There has been some work which indicates that facial expressions could be useful in the diagnosis of certain psychological disorders\cite{wang2008_neuroscience, GirardEtAl2013_MAA}. But to the best of our knowledge, until now there has been no research which establishes the relationship between facial expressions and ADHD/ASD.

In this work we aim to make the diagnostic procedure for ADHD and ASD easier, more efficient and more objective through automatic analysis of a person's behaviour. We propose a computer-vision based approach to automatically aid diagnosis of ADHD and ASD. We extract high level features from tracked faces in videos to learn classification models for ADHD and ASD prediction. We adapt a recently proposed Dynamic Deep Learning method to recognise facial action units from RGB data \cite{JaiswalValstar_DLD2016}, and use face tracking data from RGBD (colour+depth) images recorded using a Kinect 2.0 sensor camera to obtain head actions and facial animation unit parameters. 

We also present a first of its kind RGBD database in which 55 subjects who have previously been diagnosed with ADHD or ASD as well as subjects from a healthy control group were recorded in a controlled setting. We evaluate our proposed approach on this database and show that that our approach performs highly accurately ADHD and ASD classification tasks achieving classification rates of 96\% for Controls vs Condition (ADHD/ASD) group and 94\% for Comorbid (ADHD+ASD) vs ASD only group. 

In summary, our main contributions are:
\begin{itemize}
\item A novel fully automatic approach for making diagnostic predictions for ADHD and ASD directly from videos.
\item Establishing the relationship between facial expression/gestures and neurodevelopmental conditions such as like ADHD and ASD.
\item A new database for evaluating computer vision based algorithms on the task of predicting ADHD and ASD diagnosis.
\end{itemize}

\section{Related Work}
\noindent The field of using Computer vision techniques for monitoring people for ADHD and ASD is still in its infancy and there has been limited research. Below we describe some of the existing works which aim towards automatic detection of certain markers which could help in the diagnosis of ADHD and ASD.

\subsection{Detection for ADHD}
\noindent Some preliminary studies have been conducted to demonstrate the use of depth capturing cameras to monitor the activities of people. For e.g. Hernandez-Vela et al. \cite{vela2011} extracted 3D skeletal model of human body, using RGB-D image sequences. Using this skeletal model, they tracked 14 reference points corresponding to skeletal joints and used them to detect certain body gestures often found in children having ADHD. For detecting such gestures, they used Dynamic Time Warping \cite{parizeau1990}. By measuring the similarity between a temporal sequence of images with a reference sequence of a gesture, they demonstrated that they can recognize a set of defined gestures related to ADHD indicators.

In \cite{sivalingam2012}, a system was developed for tracking people across multiple cameras and sensors. They used depth measuring cameras (Microsoft Kinect) to monitor the movement of children in a classroom setting. The authors used agglomerative hierarchical clustering to segment different objects and tracked different individuals using covariance descriptors.  One of the applications they proposed for such a system would be to record the motion tracks and velocity profiles of people, to measure their activity level.

QbTest \cite{qb} is one of the most successful commercially available systems for monitoring and diagnosis of ADHD. It measures 3 main indicators of ADHD: hyperactivity, inattention and impulsivity. It combines head motion tracking with a computer based test. The head motion tracking is designed to measure the hyperactivity of the subject. For this purpose, the subject taking the test is required to wear a head band which has a reflector attached to it. The camera in front of the subject, tracks the movement of the reflector. However, the system's ability to capture the full facial information is limited as it does not track the entire face thus ignoring the 3D head pose and facial expression information. To measure the inattention and impulsivity, the subject has to take a computerised continuous performance test in which the participant has to respond quickly and accurately to certain geometrical shapes displayed on the screen. The whole test lasts for 15-20 minutes and the head motion is tracked during the entire time. After the test, the result is compared to the norm data corresponding to the subject's age and gender and a report is generated for assessment by clinicians.

\subsection{Detection for ASD}
\noindent One of the pioneering works in the field of ASD diagnosis was done by Hashemi et al. \cite{hashemi2012}. In this work, the authors developed computer vision based methods to identify certain behavioural markers based on Autism Observation Scale for Infants (AOSI) related to visual attention and motor patterns. For assessing visual attention, they focused on 3 main behavioural markers, namely sharing interest, visual tracking and disengagement of attention. These behavioural markers were detected by estimating the head pose in the left-right direction (yaw) and in the up-down direction (pitch). The head pose was estimated by tracking the position of certain facial features (eyes, nose, ear, etc.).

In \cite{rehg2011}, the authors presented another computer vision based approach for studying autism by retrieving social games and other forms of social interactions between adults and children in videos. They proposed to do this by defining social games as quasi-periodic spatio-temporal patterns. In order to retrieve such patterns from unstructured videos, the authors represent each frame using a histogram of spatio-temporal words derived from space-time interest points. The frames are clustered based on their histograms to represent the video as a sequence of cluster (keyframes) labels. The quasi-periodic pattern is found by searching for co-occurrences of these keyframe labels in time.

In \cite{rajagopalan2014}, the authors proposed an algorithm for detecting self-stimulatory behaviour which is a common behavioural marker in individuals with autism. They computed motion descriptor using dominant motion flow in the tracked body regions, to build a model for detecting self-stimulatory behaviour in videos. Similarly, in \cite{rajagopalan2015}, the authors measure children's engagement level in social interactions using low level optical flow based features.

Most of the above mentioned works have concentrated on detecting certain pre-defined behavioural markers which are often associated with either ADHD or ASD in children. They are preliminary works whose effectiveness in predicting the actual ADHD and ASD diagnosis still remains to be seen. On the other hand, this work poses the diagnosis of ADHD and ASD, directly as a machine learning problem. Our work is one of the first which attempts to learn models for directly predicting conditions such as ADHD and ASD using high level facial features which can be reliably computed nowadays. This work also differs from other works in the sense that it is mainly focusses on ADHD and ASD diagnosis in adults rather than children.

\section{Data Collection}
\label{sec:dataset}
\noindent We collected a dataset 'KOMAA' (Kinect Data for Objective Measurement of ADHD and ASD) for the purpose of evaluating our proposed method. The database consists of video recordings from a total of 57 subjects. The length of each video is approximately 12 min. and is recorded using Kinect 2.0 device which is capable of capturing high resolutions RGB and depth images. All the participants in the recording were adults over the age of 18 years. During the recordings the subjects sit in front of a computer screen and have to read and listen to a set of 12 short stories. Each story is accompanied by 2-3 questions which the subjects have to answer in their voice. These stories have been selected from the 'Strange Stories' task \cite{happe1994} which is often used as psychological test for the diagnosis of ASD. The text of each story along with the corresponding questions were displayed on the screen. Additionally, a pre-recorded voice was played reading out the story and the corresponding questions. Such a setup was prepared so as to simulate the effect of an actual person telling the story and asking the questions, but at the same time keeping the setup as automated as possible.

The subjects in this database can be divided into four different categories. The first category is the control group which consists of subjects who show no symptoms of ADHD/ASD and have never been diagnosed with either ADHD or ASD. In order to make sure that the subjects recruited in the control group do not have any chance of having ADHD or ASD, each subject was asked to complete 2 screening questionnaires: Adult ADHD Self-Report Scale (ASRS) \cite{kessler2005}) and Autism Spectrum Quotient (AQ10) \cite{baron2001}. The ASRS is a screening measure for ADHD symptoms in adults consisting of 18 items. It is based on DSM-IV items for ADHD and is considered to have excellent reliability and validity \cite{kessler2007}. Similarly, AQ10 is a screening measure for Autism symptoms consisting of 10 items and is widely used to measure the degree to which an adult has autistic traits. Only those participants who scored less than a certain threshold value in each of the questionnaires, were selected as a part of the control group. The threshold values for ASRS (Part A) and AQ10 were set to be 4 and 6 respectively. The other three categories include the ASD group (consisting of subjects who have been diagnosed with ASD), ADHD group (subjects who have been diagnosed with ADHD) and ASD+ADHD group (subjects who have been diagnosed with both ADHD and ASD. 

\begin{figure}[t]
\begin{center}
   \includegraphics[width=0.8\linewidth]{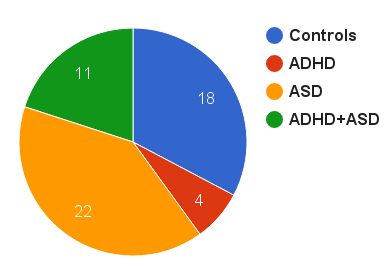}
\end{center}
   \caption{Distribution of participants in KOMAA dataset.}
\label{distribution}
\end{figure}

The total number of subjects recruited into each category is shown in Fig. \ref{distribution}.

\section{Methodology}
\noindent Training statistical machine learning based classifiers which can automatically differentiate between subjects with ADHD/ASD from healthy controls, is a difficult problem. The problem becomes even more challenging when the number of training examples are small. Deep learning based approaches which directly use low level pixel information to learn high level semantics, currently provide state-of-the-art performance on a number of computer vision tasks. However, using low level information on the limited number of training examples in our case, can lead to severe overfitting.

Our approach to training the classifiers involves computing high level feature descriptors corresponding to facial expressions (facial AUs), head pose and motion, etc. To compute the feature descriptors, each video is first divided into 12 segments corresponding to the 12 stories that the participants have to read while they were recorded. This has been done manually, but could easily be automated given that the timing of the delivery of the stories is controlled by the researcher. 

For each video segment, histogram based feature descriptors are computed separately using pre-trained classifiers/regressors that detect individual behavioural cues. Grouping these cues per story helps to preserve temporal information which would otherwise be lost if histograms would have been computed over all the frames in a video, at a small price of multiplying the dimensionality of our overall feature vector by a factor 12. The combined set of feature descriptors from all segments in a recording are used for used for training the ADHD/ASD classification models (See Fig. \ref{graphical_abstract}. Below we describe the main components of our approach in more detail.

\subsection{Feature descriptors}
\noindent Six different sets of features are computed from the recorded video of each subject. Most of the features are computed on a per-frame basis, which are then converted into multiple histograms where each histogram is computed over all the frames in a video segment. The feature descriptors used in our approach are described below:\\

\begin{figure}[t]
\begin{center}
  \includegraphics[width=1\linewidth]{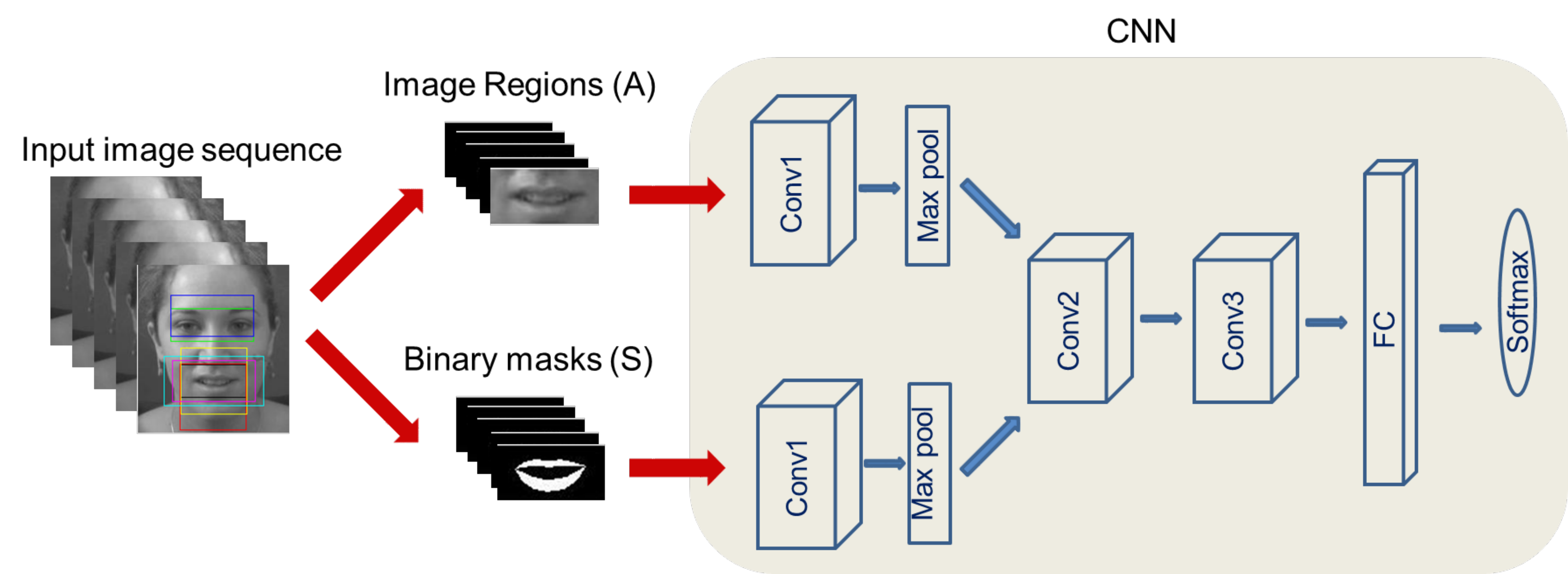}
\end{center}
  \caption{Graphical overview of the CNN based approach used for predicting facial AUs \cite{JaiswalValstar_DLD2016}.}
\label{deepAU}
\end{figure}

\noindent\textbf{1) Dynamic Deep Learned Facial Action Units:}\\ 
Facial action units (AU) are movement of individual or group of facial muscles defined according to the Facial Action Coding System (FACS)\cite{Ekman02-FAC}. Anatomically based descriptors of facial expressions, they can be a good representative of the emotional and mental state of a person and can encode a large number of social signals. Intensities for a set of 6 AUs (AU1, AU2, AU4, AU12, AU15, AU20) and occurrence for AU45 (blinks) were estimated for each frame in video. For this purpose we used AU models trained using a slightly modified version of the deep CNNs described in \cite{JaiswalValstar_DLD2016}. The network architecture used for this purpose is shown in Fig. \ref{deepAU}. This network does not use Bi-directional Long Short Term Memory (BLSTM) used by the original work in \cite{JaiswalValstar_DLD2016}. Histograms of AU intensities was computed over all the frames in a video segment. One histogram was computed for each AU consisting of 10 bins each. For AU45, the frequency of its occurrence and the average duration of its activation were estimated in each video segment. The histograms of all AU intensities and the AU45 statistics were concatenated together resulting in a 62 dimensional AU vector $F_{au}$, for each video segment.\\

\noindent\textbf{2) Kinect Animation Units:}\\
The Kinect also provides Animation Units (AnUs), geometry-based descriptors similar to mpeg-4 face animation parameters (FAPs)\cite{Pandzic2003}. While they are not based on muscle actions and can not detect facial actions that only cause appearance changes, the fact that they are obtained from RGBD data makes them very reliable. The intensity of a number of AnUs were estimated for each frame in the video using the Kinect v2 library. In order to aggregate the statistics over each video segment, a histogram of ANU intensities was computed for each facial AnU. Each histogram consisted of 10 bins resulting in a 10 dimensional feature vector corresponding to each ANU. A total of 12 AnUs (6 corresponding to left and 6 to right part of the face) were used. The histograms from all 12 AnUs were concatenated, resulting in 120 dimensional AnU vector $F_{an}$, for each video segment.\\

\noindent\textbf{3) Head Pose:}\\
One of the major challenges for people with ADHD is their inability to do tasks which requires sustained attention. The pose of the head (in 3D space) can provide valuable cues about the attention state of a person at a certain instance of time. Since the participants in our study were required to complete the task by looking the computer screen, any deviation of the head pose away from the computer screen would indicate loss of attention.

\begin{figure}[h]
\begin{center}
   \includegraphics[width=0.4\linewidth]{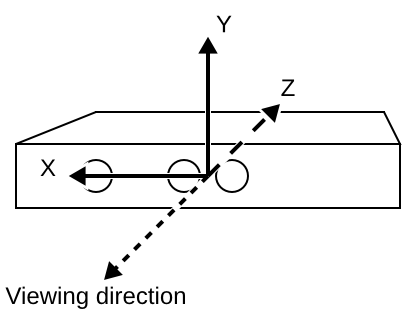}
\end{center}
   \caption{Kinect coordinate system.}
\label{kinect_coordinates}
\end{figure}

The rotation of the head about the X, Y and Z axis (pitch, yaw and roll) were estimated for each frame of the video using the Kinect v2 software. The X, Y and Z axis are defined in reference to the location of the Kinect device as shown in Fig. \ref{kinect_coordinates}. We assumed the median pose of the head to be the most attentive state. Rotation of the head away from the median pose were computed about the X, Y and Z axis separately. Histograms of these rotation angles were computed over the video segments for each axis separately. Each histogram consisted of 18 bins with equally spaced bin centres ranging from -45 to 45. This resulted in a 54 dimensional head pose vector $F_{hp}$, for each video segment.\\  

\noindent\textbf{4) Speed of head movement:}\\
Dynamics of head motion has been a less researched aspect in the field of psychological disorders. In order to investigate the role head motion, we estimated the speed of head motion at each frame of the video. For this purpose, we selected a set of stable facial landmarks belonging to eye corners and 4 points on the nose. The location of these stable facial landmarks are invariant to changes in facial expressions and hence suitable for estimating the motion of the head. The motion of the head is estimated by computing the location of the centroid $C_i$ of the stable landmarks. The speed of head motion $S_i$ at any frame $i$ can be estimated by computing the displacement of the centroid as given below:
\begin{equation}
S_i = ||C_i-C_{i-1}||*f
\end{equation}
where $f$ is the frame rate of the recorded video. In order to make speed estimation more reliable and invariant to any fluctuations in the frame rate, the estimated speed was smoothed by computing a moving average over 20 consecutive frames.

A histogram of the estimated speeds was computed to aggregate the statistics over each video segment. The histograms consists of 10 bins resulting in a 10 dimensional speed vector $F_{sp}$, for each video segment.\\

\noindent\textbf{5) Cumulative Distance:}\\
Hyperactivity is another major challenge associated with ADHD, implying that individuals with ADHD tend to display much higher levels of motoric behaviour than healthy individuals. The movement can be in the form of whole body movement or smaller movements confined to head (rotation) or hands and legs (fidgeting). To encode such information, the cumulative distance $F_{cd}$ moved by the head during an entire video segment, was estimated by summing up the displacements of the centroid $C_i$ given below:
\begin{equation}
F_{cd} = \sum_{i=1}^n ||C_i-C_{i-1}||
\end{equation}
where n is the total number of frames in the video segment.\\

\noindent\textbf{6) Response Times:}\\
The time taken to respond to each set of questions in the study was also used as features. Since there were 12 stories, each comprising a set of questions, a 12 dimensional response time vector $F_{rt}$ was defined consisting of the response times (in seconds) for each set of questions.\\

\subsection{Feature pre-processing and training models}

\noindent\textbf{Normalization:}
Each set of features (except the $F_{rt}$) were divided by the total number of frames in the video segment, to make them invariant to the length of video recording. The final set of features $F$ was obtained by concatenating all sets of features ($F_{au}$, $F_{an}$, $F_{hp}$, $F_{sp}$, $F_{cd}$, $F_{rt}$) from all video segments.
Each dimension in the resulting feature vector $F$ is further normalized by computing the Z-scores.

\noindent\textbf{Feature selection and training models}
Due to the high dimensionality of the resulting feature $F$ compared to the number of training examples, any classifier trained directly on the entire feature-set is most likely to overfit the training data. In order to avoid such problem, a greedy forward feature selection was employed to capture the most relevant features and reduce the dimensionality. The classification models were trained using Support Vector Machines (SVM) with a Radial Basis function kernel.

\section{Experiments and Discussion}

\noindent Our approach was evaluated on the KOMAA dataset that we collected for this purpose from a total of 57 participants (see section \ref{sec:dataset}). The distribution of participants with ADHD, ASD and healthy controls is shown in Fig. \ref{distribution}. 

To evaluate the performance of our approach in classifying each subject to the ASD, ADHD or the Control group, we followed a a 2 step procedure: In the first step we trained a classifier to distinguish between controls and condition group (participants diagnosed with either ADHD, ASD or both). In the second step, we trained another classifier to distinguish between ASD only group and Comorbid (ASD+ADHD) group. Since the ADHD only group was too small (only 4 participants), we did not had enough data to learn a robust classifier for this group.

\begin{table}[h]
\caption{Classification results for Controls vs Condition (ASD/ADHD) group.}
\label{class_results1}
\begin{center}
\begin{tabular}{ c | l | l }
  \hline
	Classifier & Correct & Incorrect\\
	\hline
  Controls  & 16 & 2\\
  Condition & 37 & 0\\
	\hline
\end{tabular}
\end{center}
\end{table}

\begin{table}[h]
\caption{Classification results for Comorbid (ADHD+ASD) vs ASD group.}
\label{class_results1}
\begin{center}
\begin{tabular}{ c | l | l }
  \hline
	Classifier & Correct & Incorrect\\
	\hline
  Comorbid  & 9 & 2\\
  ASD only & 22 & 0\\
	\hline
\end{tabular}
\end{center}
\end{table}

Our approach was evaluated using a leave-one-subject-out protocol, in which one subject is used for testing and the rest of the subjects are used for training. This process is repeated for each subject and the overall score is obtained by averaging over each test subject. The classification performance of our approach is shown in Table \ref{class_results1} and \ref{class_results2}. For classification into Control and Condition group, we obtain a very high classification accuracy of 96.4\%. Similarly, for classification into Comorbid(ASD+ADHD) and ASD only group, we obtain a high classification accuracy of 93.9\%. 

\begin{figure}[t]
\begin{center}
   \includegraphics[width=0.8\linewidth]{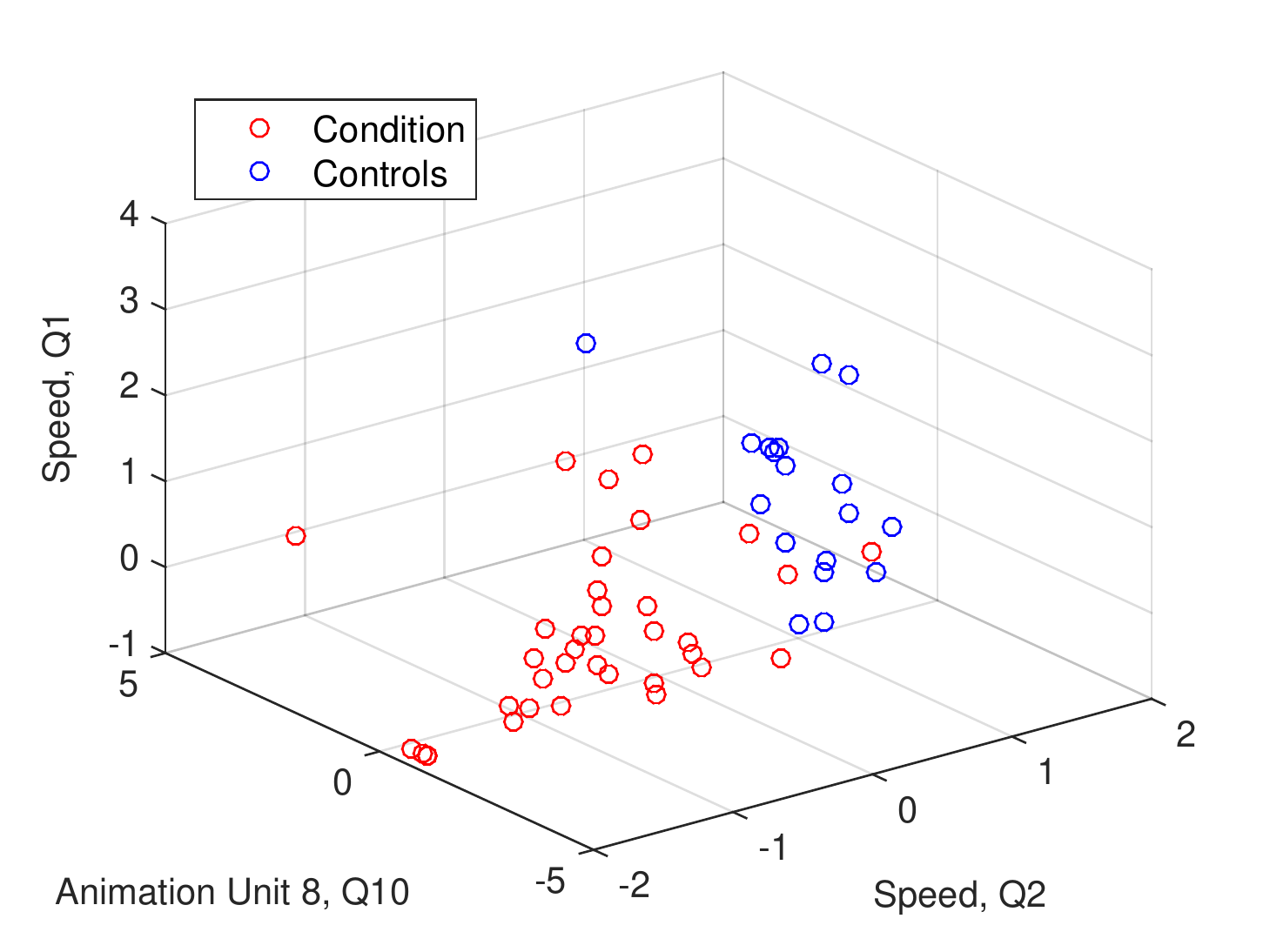}
\end{center}
   \caption{Top 3 features distinguishing Condition (ASD/ADHD) from control group.}
\label{plot_condition}
\end{figure}

\begin{figure}[t]
\begin{center}
   \includegraphics[width=0.8\linewidth]{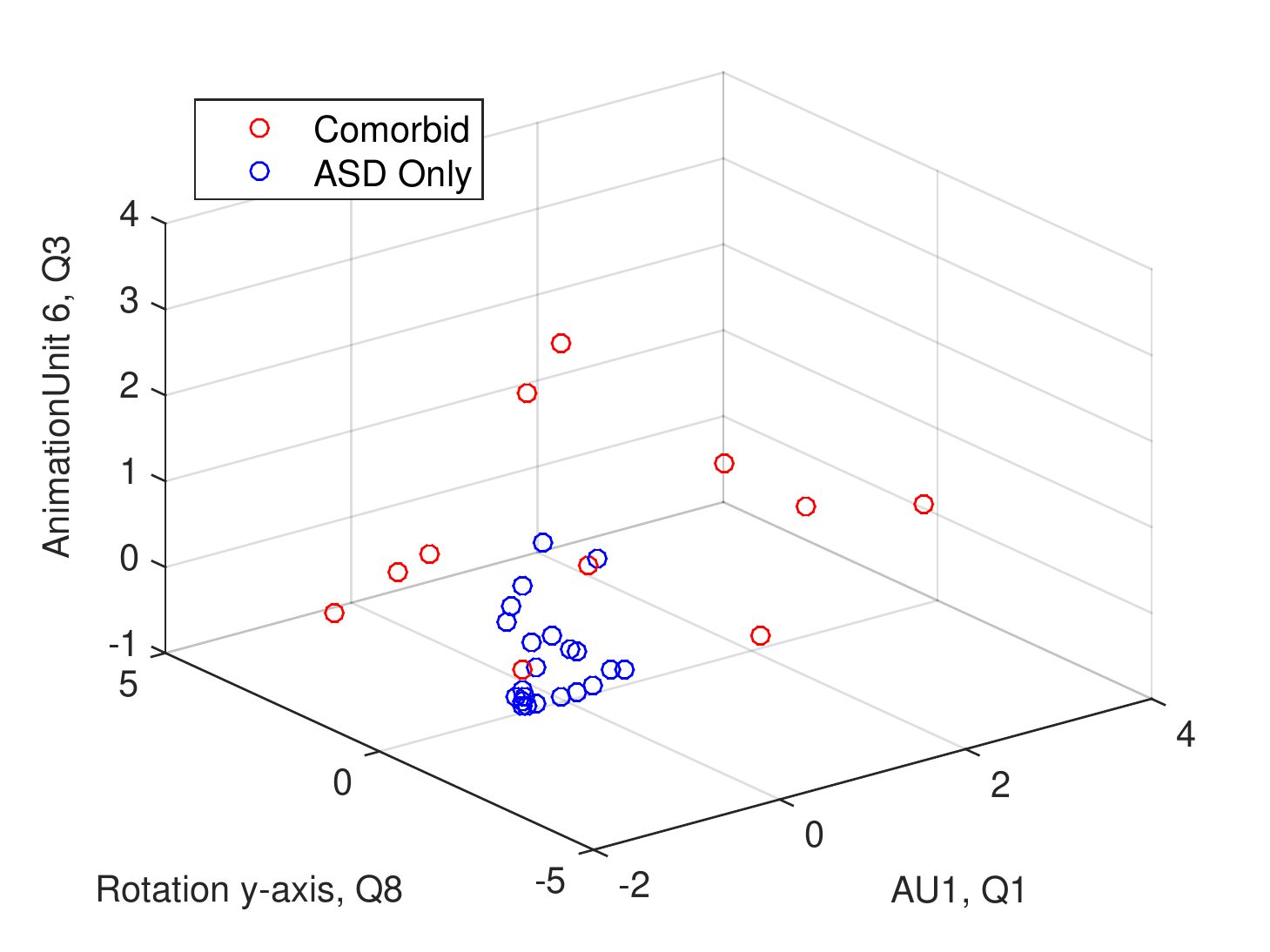}
\end{center}
   \caption{Top 3 features distinguishing Comorbid (ASD+ADHD) from ASD only group.}
\label{plot_comorbid}
\end{figure}

Looking at the individual contribution of different cues, Fig. \ref{plot_condition} and \ref{plot_comorbid}, show the class separation provided by some of the important features selected by using the forward feature selection approach. From these figures, we can observe that for classification of Controls and Condition group, features such as Speed of head motion (from video segment corresponding to Question 1 and 2) and Animation Unit 8 (from video segment corresponding to Question 10 of 'Stange Stories task') were found to be most discriminative. For Comorbid vs ASD classification, AU1 (inner-brow raiser), AnU6 (lip-corner puller) and head rotation about Y-axis turn out be highly discriminative These features were extracted from the video segment corresponding to Question 1, 3 and 8 of the Strange stories task respectively.

\section{Conclusions}
We presented a novel method for making diagnostic prediction of ADHD and ASD in test subjects through automatic video analysis. Facial cues such as head motion, facial expression and pose are used in learning  models which can accurately predict ADHD and ASD. The role of facial expressions as a potential feature for classification of individuals with these disorders from healthy controls, was investigated. A high performance was achieved in terms of classification accuracy, which indicates a high potential for facial expressions and other facial gestures to be used for making automatic predictions for ADHD, ASD and other neurodevelopmental disorders.

\bibliographystyle{unsrt}
\bibliography{paper}

\end{document}